\documentclass{article}
\usepackage[nonatbib,preprint]{neurips_2021}
\usepackage[numbers]{natbib}
\usepackage[utf8]{inputenc} 
\usepackage[T1]{fontenc}    
\usepackage{hyperref}       
\usepackage{url}            
\usepackage{booktabs}       
\usepackage{amsfonts}       
\usepackage{nicefrac}       
\usepackage{microtype}      
\usepackage[export]{adjustbox}

\usepackage{mdframed}
\usepackage{amsmath,mathrsfs,amssymb,mathtools,bm}
\usepackage{scalerel}
\usepackage{amsthm}
\usepackage{txfonts}
\usepackage{fontawesome}
\usepackage{wrapfig}
\usepackage{todonotes}
\usepackage{xcolor}
\usepackage{setspace}
\usepackage{multirow}
\usepackage{bm,bbm}
\usepackage{paralist}
\usepackage{enumitem}
\usepackage{caption}[small]
\usepackage{subcaption} 

\newtheorem{problem*}{Problem}

\newtheorem*{example*}{Example}
\usepackage[ruled,noresetcount,vlined]{algorithm2e}
\makeatletter

\newcommand{\nosemic}{\renewcommand{\@endalgocfline}{\relax}}
\newcommand{\dosemic}{\renewcommand{\@endalgocfline}{\algocf@endline}}
\let\oldnl\nl
\newcommand{\nonl}{\renewcommand{\nl}{\let\nl\oldnl}}

\SetKwFor{Repeat}{repeat}{:}{endw}
\makeatother

\usepackage{todonotes}
\usepackage{xcolor}
\usepackage{setspace}
\usepackage{bm,bbm}
\usepackage{paralist}

\DeclareMathOperator*{\argmax}{argmax}
\DeclareMathOperator*{\argmin}{argmin}
\DeclareMathOperator*{\minimize}{minimize}

\usepackage{multicol}
\usepackage{float}
\floatstyle{ruled}
\newfloat{model}{thp}{lop}
\floatname{model}{Model}

\newcommand{\jssdnn}{\textsl{JSS-DNN}}

\newcommand{\rev}[1]{{\color{purple}{#1}}}

\newcommand{\cI}{\mathcal{I}}

 \newcommand{\RR}{\mathbb{R}}

\newcommand{\btheta}{{\bm{\theta}}}

\newcommand\hugP[1]{\left(#1\right)}

\title{Fast Approximations for Job Shop Scheduling:\\ 
A Lagrangian Dual Deep Learning Method}

\author{%
James Kotary\\
{Syracuse University}\\
\texttt{jkotary@syr.edu}
\And
Ferdinando Fioretto\\
{Syracuse University}\\ 
\texttt{ffiorett@syr.edu}
\And
Pascal Van Hentenryck\\
{Georgia Institute of Technology}\\
\texttt{pvh@isye.gatech.edu}
}

\begin{document}

\maketitle\allowdisplaybreaks\sloppy

\begin{abstract}
The Jobs shop Scheduling Problem (JSP) is a canonical combinatorial 
optimization problem that is routinely solved for a variety of industrial 
purposes. It models the optimal scheduling of multiple sequences of 
tasks, each under a fixed order of operations, in which individual 
tasks require exclusive access to a predetermined resource for a 
specified processing time. The problem is NP-hard and computationally 
challenging even for medium-sized instances. Motivated by the increased 
stochasticity in production chains, this paper explores a deep learning 
approach to deliver efficient and accurate approximations to the 
JSP. In particular, this paper proposes the design of a deep 
neural network architecture to exploit the problem structure, its 
integration with Lagrangian duality to capture the problem constraints, 
and a post-processing optimization to guarantee solution feasibility.
The resulting method, called \textsl{JSP-DNN}, is evaluated on hard JSP 
instances from the JSPLIB benchmark library. Computational results 
show that \textsl{JSP-DNN} can produce JSP approximations of high quality at 
negligible computational costs.
\end{abstract}


\section{Introduction}

The Job shop Scheduling Problem (JSP) is defined in terms of a set of jobs, each of which consists of a sequence of tasks. Each task is processed on a predetermined resource and no two tasks can overlap in time on these resources. The goal of the JSP is to sequence the tasks in order to minimize the total duration of the schedule. Although the problem is NP-hard and computationally challenging even for medium-sized instances, it constitutes a fundamental building block for the optimization of many industrial processes and is key to the stability of their operations. Its effects are profound in our society, with applications ranging from supply chains and logistics, 
to employees rostering, marketing campaigns, and manufacturing to name just a few \cite{kan2012machine}.

While the Artificial Intelligence and Operations Research communities have contributed fundamental advances in optimization in recent decades, the complexity of these problems often prevents them from being effectively adopted in contexts where many instances must be solved over a long-term horizon (e.g., multi-year planning studies) or when solutions must be produced under stringent time constraints.
For example, when a malfunction occurs or when operating conditions 
require a new schedule, replanning needs to be executed promptly as 
machine idle time can be extremely costly. (e.g., on the order of 
\$10,000 per minute for some applications \cite{gombolay2018human}). 
To address this issue, system operators typically seek approximate solutions to the original scheduling problems. However, while more efficient computationally, their sub-optimality may induce substantial economical and societal losses, or they may even fail to satisfy important constraints.

Fortunately, in many practical settings, one is interested in solving 
many instances sharing similar patterns.
Therefore, the application of deep learning methods to aid the resolution of these optimization problems is gaining traction in the nascent area at the intersection between constrained optimization and machine 
learning \cite{bengio2020machine,kotary2021end,vesselinova2020learning}.
In particular, supervised learning frameworks can train a model using 
pre-solved optimization instances and their solutions. 
{However, while much of the recent progress at the intersection of 
constrained optimization and machine learning has focused on learning 
good approximations by jointly training prediction and 
optimization models \cite{balcan2018learning,khalil2016learning,nair2020solving,nowak2018revised,vinyals2017pointer} 
and incorporating optimization algorithms into differentiable 
systems \cite{amos2019optnet,vlastelica2020differentiation,wilder2018melding,mandi2019smart}, 
learning the combinatorial structure of complex optimization problems 
remains a difficult task.}
In this context, the JSP is particularly challenging 
due to the presence of disjunctive constraints, which present some unique 
challenges for machine learning. 
Ignoring these constraints produce unreliable and unusable approximations, 
as illustrated in Section \rev{\ref{sec:challenges}}.

JSP instances typically vary along two 
main sets of parameters: {\bf (1)} the continuous \emph{task durations} 
and {\bf (2)} the combinatorial \emph{machine assignments} associated with each task. This work focuses on the former aspect, addressing the problem of learning 
to map JSP instances from a distribution over task durations to solution 
schedules which are close to optimal. Within this scope, the desired mapping 
is combinatorial in its structure: a marginal increase in one task duration 
can have cascading effects on the scheduling system, leading to significant 
reordering of tasks between respective optimal schedules \cite{kan2012machine}. 

To this end, this paper integrates Lagrangian duality within a Deep Learning framework to ``enforce'' constraints  when learning job shop schedules. Its key idea is to exploit Lagrangian duality, which is widely used to obtain tight bounds in optimization, during the training cycle of a deep learning model. The paper also
proposes a dedicated deep-learning architecture that exploits the structure of JSP problems and an efficient post-processing step to restore feasibility of the predicted schedules. 

\smallskip\noindent\textbf{Contributions}
The contributions of this paper can be summarized as follows.  
\textbf{(1)} It proposes \jssdnn, an approach that uses a deep neural 
network to accurately predict the tasks start times for the JSP.
\textbf{(2)} \jssdnn{} captures the JSP constraints using 
a Lagrangian framework, recasting the JSP prediction problem 
as the Lagrangian dual of the constrained learning task and using a 
  subgradient method to obtain high-quality solutions. 
\textbf{(3)} It further exploits the JSP structure through the 
design of a bespoke network architecture that uses two dedicated 
sets of layers: \emph{job layers} and \emph{machine layers}. They reflect the 
task-precedence and no-overlapping structure of the JSP, respectively, 
encouraging the predictions to take account of these constraints. 
\textbf{(4)} 
While the adoption of Lagrangian duals and the dedicated JSP network 
architecture represent notable improvements to the prediction, the 
model predictions represent approximate solutions to the JSP  
and may not feasible. In order to derive feasible solutions from these 
predictions, this paper proposes an efficient reconstruction technique. 
\textbf{(5)} Finally, experiments against highly optimized industrial solvers show that 
JSP-DNN provides state-of-the-art JSP approximations, 
both in terms of accuracy and efficiency, on a variety of standard benchmarks. 
To the best of the authors' knowledge, this work is the first to tackle 
the predictions of JSPs using a dedicated supervised learning 
solution. 

\section{Related Work}
The application of Deep Learning to constrained optimization problems
is receiving increasing attention. Approaches which learn solutions to 
combinatorial optimization using neural networks include
\cite{pointer_nets,khalil2017learning,kool2018attention}. These
approaches often rely on predicting permutations or combinations as 
sequences of pointers. Another line of work leverages explicit
optimization algorithms as a differentiable layer into neural networks
\cite{amos2017optnet,donti2017task,wilder2019melding}.  
An in-depth review of these topics is provided in \cite{KotaryFHW21}.

A further collection of works interpret constrained optimization as a two-player
game, in which one player optimizes the objective function and a
second player attempt at satisfying the problem constraints
\cite{kearns2017,narasimhan2018,agarwal2018,Fioretto1}.  
For instance, \citet{agarwal2018} 
proposes a best-response algorithm applied to
fair classification for a class of linear fairness constraints. To
study generalization performance of training algorithms that learn to
satisfy the problem constraints, ~\citet{cotter2018}
propose a two-players game approach in which one player optimizes the
model parameters on the training data and the other player optimizes
the constraints on a validation set. 
\citet{arora2012multiplicative} propose the use of a multiplicative rule 
to maintain some properties by iteratively changing the weights of 
different distributions; they also discuss the applicability of the 
approach to a constraint satisfaction domain. 

Different from these proposals, this paper proposes a framework that
exploits key ideas from Lagrangian duality to encourage the satisfaction 
of generic constraints within a neural network learning cycle and apply 
them to solve complex JSP instances. 

This paper builds on the recent results that were dedicated to learning 
and optimization in power systems \cite{FiorettoMH20,FiorettoHM0B020}.

\section{Preliminaries}
\label{sec:preliminaries}

\subsection{Job Shop Scheduling Problem}
The JSP is a combinatorial optimization problem in which $J$ jobs, each composed of $T$ tasks, must be processed on $M$ machines. 
Each job comprises a sequence of $T$ tasks, each of which is assigned
to a different machine. Tasks within a job must be processed in
their specified sequential order. Moreover, no two tasks may occupy the same 
machine at the same time. The objective is to find a schedule which minimizes the time to process all tasks, known as the \emph{makespan}. This paper considers the classical setting in which the number of tasks in each job is equal to the number of machines ($T=M$), so that  each job has one task assigned to each machine. This leads to a problem size $n = J \times M$. This is however not a limitation of the proposed work and its implementation generalizes beyond this setting.

\begin{model}[!t]
{
\caption{JSP Problem}
\label{model:jss}
\vspace{-6pt}
\begin{align}
    \mathcal{P}(\bm{d}) = 
    \textstyle\argmin_{\bm{s}}\;\;& u
    \label{load_flow_obj} \\
    \mbox{subject to:}\;\;
        & u \geq s_T^j + d_T^j                  \;\; \forall j \!\in\! [J]                             \label{con:2a} \tag{2a}\\
        & s_{t+1}^j \geq s_t^j + {d}^{j}_{t}    \;\; \forall j \!\in\! [J-1], \forall t \!\in\! [T]    \label{con:2b} \tag{2b}\\
        & s_t^{j} \geq s_{t'}^{j'} + {d}_{t'}^{j'}\; \lor s_{t'}^{j'} \geq s_{t}^{j} + {d}_{t}^{j}  \label{con:2c} \tag{2c}\\
        & \;\;\qquad \forall j, j' \!\in\! [J]: j \neq j', t,t' \!\in\! [T] \,\text{with}\, {\sigma}_{t}^{j} = {\sigma}_{t'}^{j'}  \notag \\
        & s_{t}^{j} \!\in\! \mathbb{N}          \;\; \forall j \!\in\! [J], t \!\in\! [T] \label{con:2d} \tag{2d}
    \end{align}
    }
    \vspace{-12pt}
\end{model}

The optimization problem associated with a  JSP instance 
is described in Model \ref{model:jss}, where
${\sigma}_{t}^{j}$ denotes the machine that processes task 
$t$ of job $j$, and ${d}_t^j$ denotes the processing time on
machine $\sigma_{t}^{j}$ needed to complete task $t$ of job $j$. 
The decision variables $s_t^j$
and $u$ represent, respectively, the start times of each task and the 
makespan. 
In the following, $\bm{d}$ and $\bm{s}$ denote, respectively, the input vector (processing times) and output vector (start times), and  $\mathcal{P}(\bm{d})$ represents the optimal solution of a JSP instance with inputs $\bm{d}$. For simplicity, we assume that this solution is unique, i.e., there is a rule to break ties when there are multiple optimal solutions. 

The \emph{task-precedence} constraints \eqref{con:2b} require that all 
tasks be processed in the specified order; the \emph{no-overlap} constraints 
\eqref{con:2c} require that no two tasks using the same machine overlap in time. 
The difficulty of the problem comes primarily from the disjunctive constraints \eqref{con:2c} defining the no-overlap condition. The JSP is, in general, NP-hard and can be formulated in various ways, including several Mixed Integer Program (MIP), and Constraint Programming (CP) models, each having distinct characteristics \cite{ku2016mixed}. In the following sections, ${\cal C}(\bm{s}, \bm{d})$ denotes
the set of constraints \eqref{con:2b}--\eqref{con:2d} associated with 
problem ${\cal P}(\bm{d})$.

In the following sections, ${\cal C}(\bm{s}, \bm{d})$ denotes
the set of constraints \eqref{con:2b}--\eqref{con:2d} associated with 
problem ${\cal P}(\bm{d})$.

\subsection{Deep Learning Models}
Supervised Deep Learning  can be viewed as the task of
approximating a complex non-linear mapping from labeled data.  Deep
Neural Networks (DNNs) are deep learning architectures composed of a
sequence of layers, each typically taking as inputs the results of the
previous layer \cite{lecun2015deep}. Feed-forward neural networks are
basic DNNs where the layers are fully connected and the function
connecting the layer is given by
\[
\bm{o} = \Gamma(\bm{W} \bm{x} + \bm{b}),
\]
where $\bm{x} \!\in\! \RR^n$ and is the input vector, $\bm{o} \!\in\! \RR^m$ the output vector, $\bm{W} \!\in\! \RR^{m \times n}$ a matrix of weights, and
$\bm{b} \!\in\! \RR^m$ a bias vector. The function $\Gamma(\cdot)$ is
often non-linear (e.g., a rectified linear unit (ReLU)). 
The matrix of weights $\bm{W}$ and the bias vector $\bm{b}$ are referred 
to as, \emph{network parameters}, and denoted with $\bm{\theta}$ in the 
remainder of this paper.

\section{JSP Learning Goals}
Given the set of processing times $\bm{d} \!=\! (d_t^j)_{j\in[J],t\in[T]}$ 
associated with each problem task 
(as well as a static assignment of tasks to machines $\bm{\sigma}$), 
the paper develops a JSP mapping $\hat{\mathcal{P}}: \mathbb{N}^n \to \mathbb{N}^n$
to predict the start times $\bm{s} \!=\! (s_t^j)_{j\in[J],t\in[T]}$ for each task.  
The input of the learning task is a dataset ${\cal D} =
\{(\bm{d}_i, \bm{s}_i)\}_{i\!=\!1}^N$, where $\bm{d}_i$
and $\bm{s}_i$ represent the $i^{th}$ instance of task
processing times and start times that satisfy $\bm{s}_i \!=\!
\mathcal{P}(\bm{d}_i)$.  The output is a JSP approximation 
function $\hat{\cal P}_{\bm{\theta}}$, parametrized by vector $\bm{\theta} \in \RR^k$, 
that ideally would be the result of the following optimization 
problem 
\begin{align*}
\minimize_{\bm{\theta}} & \;\; \sum_{i=1}^N 
    {\cal L}\left(\bm{s}_i,\hat{\cal P}_{\bm{\theta}}(\bm{d}_i)\right) \;\;
    \mbox{subject to:} \;\; {\cal C}\left(\hat{\cal P}_{\bm{\theta}}(\bm{d}_i), \bm{d}_i\right),
\end{align*}
\noindent
whose loss function ${\cal L}$ captures the prediction accuracy of model 
$\hat{{\cal P}}_{\bm{\theta}}$ 
and ${\cal C}(\hat{\bm{s}},\bm{d})$ holds if the predicted start 
times $\hat{\bm{s}} = \hat{{\cal P}}_{\bm{\theta}}(\bm{d})$ produce a feasible solution to the JSP constraints.

One of the key difficulties of this learning task is the presence of
the combinatorial feasibility constraints in the JSP. The
approximation $\hat{\cal P}_{\bm{\theta}}$ will typically not satisfy 
the problem constraints, as shown in the next section. 
After exposing this challenge, this paper combines three techniques to obtain a feasible solution:
\begin{enumerate}[leftmargin=*, parsep=0pt, itemsep=0pt, topsep=0pt]
\item it learns predictions which are near-feasible using an augmented loss function;
\item it exploits the JSP structure through the design of a neural network architecture, and
\item it efficiently transforms these predictions into nearby solutions that satisfy the JSP constraints.  
\end{enumerate}  

\smallskip\noindent
Combined, these techniques form a framework for predicting accurate and feasible JSP scheduling approximations.

\section{The Baseline Model and its Challenges}
\label{sec:challenges}

\begin{figure}[!t]
\centering
  \includegraphics[width=0.38\linewidth,valign=c]{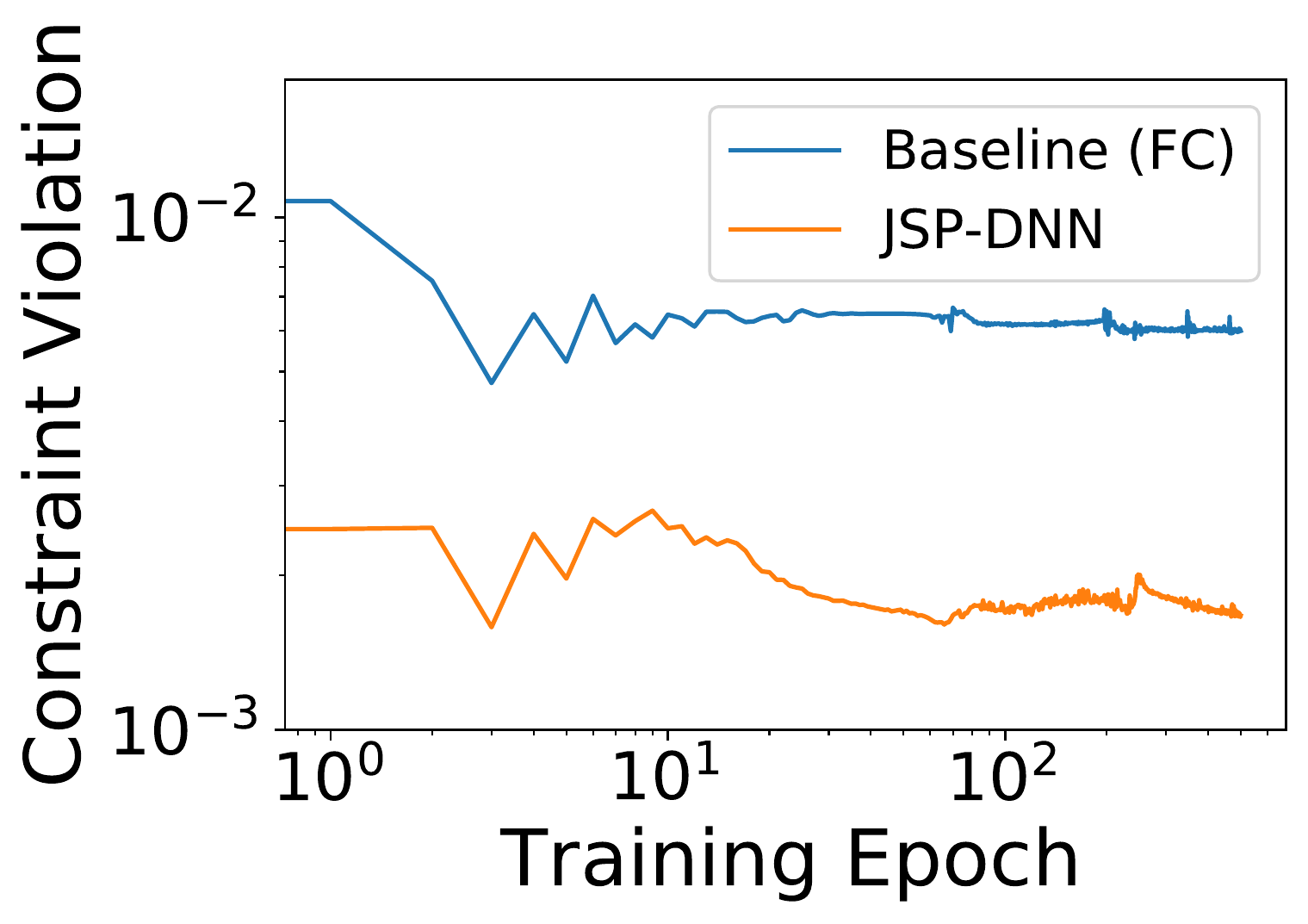} \hspace{12pt}
  \includegraphics[width=0.38\linewidth,valign=c]{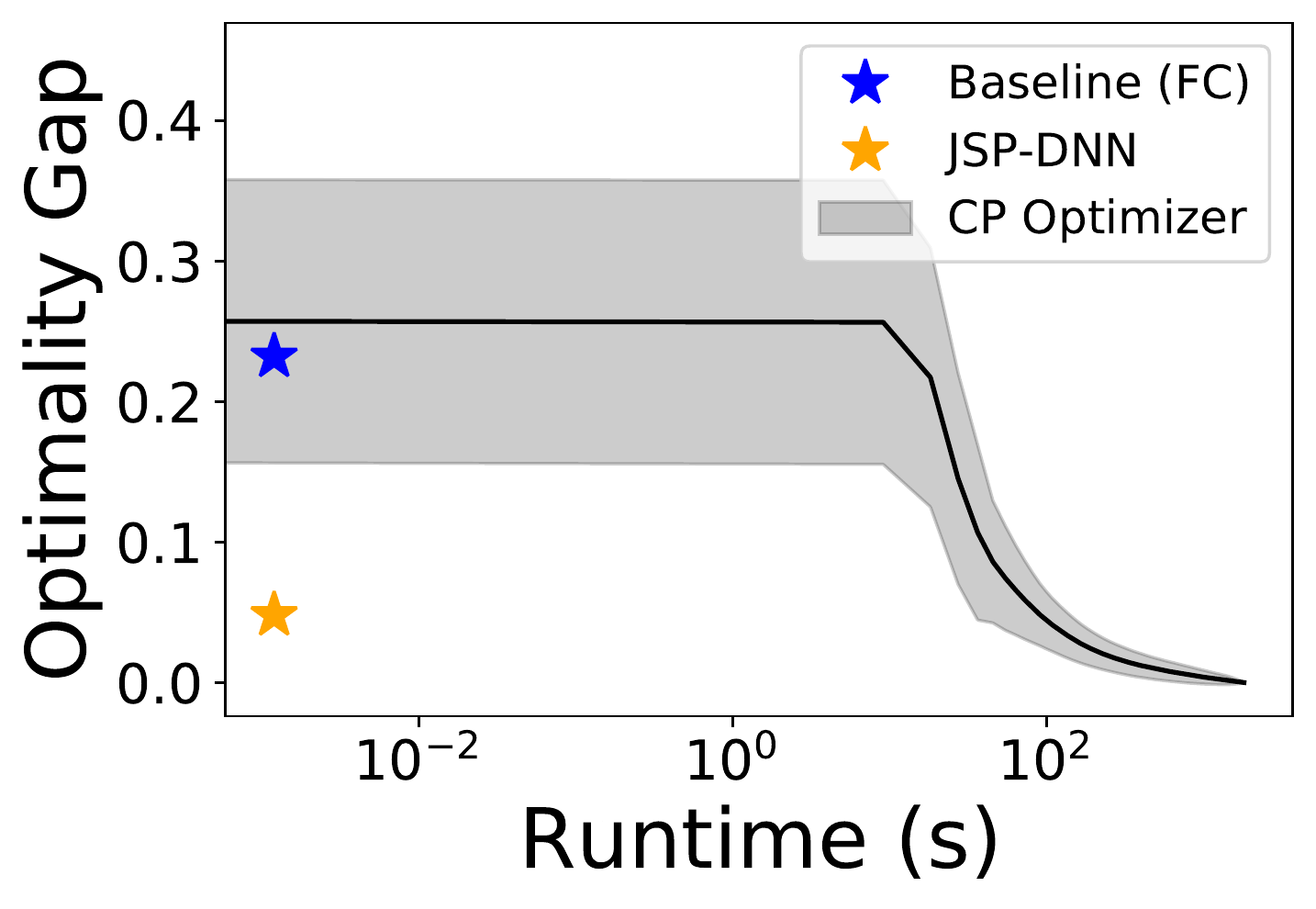}
  \caption{Constraint violations (left) and performance (right) of the 
  baseline model (fully connected network trained with MSE loss) compared to the proposed JSP-DNN model. Benchmark: \textsl{swv11}. 
  Constraint violation is the average magnitude of task overlap, measured 
  with respect to the average processing time. 
  The performance (right) reports the relative difference in makespan attained by the baseline and the JSP-DNN models, compared to an optimized 
  version of IBM CP-optimizer over $30$ minutes.}
  \label{fig:motivation_1}
\end{figure}

This paper first presents the results of a baseline model whose approximation $\hat{\cal P}_{\bm{\theta}}$ is learned from a feed-forward ReLU network, named {$\textsl{FC}$}. The challenges of FC are
illustrated in Figure \ref{fig:motivation_1}, which reports the constraint violations (left) and the performance (right) of this baseline model compared to the proposed \textsl{JSP-DNN} model on the \textsl{swv11} JSP benchmark instance of size $50 \times 10$ (see Section \ref{sec:experiments} for details about the models and datasets). The left figure reports the non-overlap constraint violations measured as the relative task overlap with respect to the average processing time. 
While the baseline model often reports predictions whose corresponding
makespans are close to the ground truth, the authors have observed that 
this model learns to ``squeeze'' schedules in order to 
reduce their makespans by allowing overlapping tasks. As a result this baseline model converges to solutions that violate the non-overlap constraint, often 
by very large amounts. The performance plot (right) shed additional light on the usability of this model in practice.  It reports the time required by this
baseline model to obtain a feasible solution (blue star) against the solution quality of a highly optimized solver (IBM CP-optimizer) over time. Obtaining feasible solutions efficiently is discussed in Section \ref{sec:projections}. The figures also report the constraint violations and solution quality
found by the proposed \jssdnn{} model (orange colors), which show dramatic 
improvements on both metrics. The next sections present the characteristics of \jssdnn{}.

\section{Capturing the JSS Constraints}
To capture the JSS constraints within a deep learning model,  a Lagrangian relaxation approach is used.
Consider the optimization problem
\begin{equation}
\label{eq:problem}
  {\cal P} = \argmin_{y} f(y) \;\;
  \mbox{subject to } \;\; g_i(y) \leq 0 \;\; (\forall i \in [m]).
\end{equation}
In \emph{Lagrangian relaxation}, some or all the problem constraints
are relaxed into the objective function using \emph{Lagrangian
  multipliers} to capture the penalty induced by violating them. 
  When all the constraints are relaxed, the \emph{Lagrangian
  function} becomes
\begin{equation}
\label{eq:lagrangian_function}
f_{\bm{\lambda}}(y) = f(y) + \sum_{i=1}^m \lambda_i g_i(y),
\end{equation}
where the terms $\lambda_i \geq 0$ describe the Lagrangian multipliers, 
and $\bm{\lambda} = (\lambda_1, \ldots, \lambda_m)$ denotes the vector 
of all multipliers associated to the problem constraints.
Note that, in this formulation, $g(y)$ can be positive or negative. 
An alternative formulation, used in augmented Lagrangian methods 
\cite{Hestenes:69} and constraint programming \cite{Fontaine:14}, 
uses the following Lagrangian function 
\begin{equation}
\label{eq:lagrangian_function2}
f_{\bm{\lambda}}(y) = f(y) + \sum_{i=1}^m \lambda_i \max(0, g_i(y)),
\end{equation}
where the expressions $\max(0, g_i(y))$ capture a quantification of the constraint violations. 
%
When using a Lagrangian function, the optimization problem becomes
\begin{equation}
\label{eq:LR}
LR_{\bm{\lambda}} = \argmin_y f_{\bm{\lambda}}(y),
\end{equation}
and it satisfies $f(LR_{\bm{\lambda}}) \leq f({\cal P})$. That is, 
the Lagrangian function is a lower bound for the original function. 
Finally, to obtain the strongest Lagrangian relaxation of ${\cal P}$,
the \emph{Lagrangian dual} can be used to find the best Lagrangian
multipliers,
\begin{equation}
\label{eq:LD}
LD = \argmax_{\bm{\lambda} \geq 0} f(LR_{\bm{\lambda}}).
\end{equation}
For various classes of problems, the Lagrangian dual is a strong
approximation of ${\cal P}$. Moreover, its optimal solutions can often
be translated into high-quality feasible solutions by a
post-processing step, which is the subject of Section \ref{sec:projections}.

This paper first presents the results of a baseline model whose approximation $\hat{\cal P}_{\bm{\theta}}$ is learned from a feed-forward ReLU network, named {$\textsl{FC}$}. The challenges of FC are
illustrated in Figure \ref{fig:motivation_1}, which reports the constraint violations (left) and the performance (right) of this baseline model compared to the proposed \textsl{JSP-DNN} model on the \textsl{swv11} JSP benchmark instance of size $50 \times 10$ (see Section \ref{sec:experiments} for details about the models and datasets). The left figure reports the non-overlap constraint violations measured as the relative task overlap with respect to the average processing time. 
While the baseline model often reports predictions whose corresponding
makespans are close to the ground truth, the authors have observed that 
this model learns to ``squeeze'' schedules in order to 
reduce their makespans by allowing overlapping tasks. As a result this baseline model converges to solutions that violate the non-overlap constraint, often 
by very large amounts. The performance plot (right) shed additional light on the usability of this model in practice.  It reports the time required by this
baseline model to obtain a feasible solution (blue star) against the solution quality of a highly optimized solver (IBM CP-optimizer) over time. Obtaining feasible solutions efficiently is discussed in Section \ref{sec:projections}. The figures also report the constraint violations and solution quality
found by the proposed \jssdnn{} model (orange colors), which show dramatic 
improvements on both metrics. The next sections present the characteristics of \jssdnn{}.

\section{Capturing the JSP Constraints}
To capture the JSP constraints within a deep learning model,  a Lagrangian relaxation approach is used.
Consider the optimization problem
\begin{equation}
\label{eq:problem}
  {\cal P} = \argmin_{y} f(y) \;\;
  \mbox{subject to } \;\; g_i(y) \leq 0 \;\; (\forall i \in [m]).
\end{equation}
Its \emph{Lagrangian function} is expressed as
\begin{equation}
\label{eq:lagrangian_function2}
f_{\bm{\lambda}}(y) = f(y) + \sum_{i=1}^m \lambda_i \max(0, g_i(y)),
\end{equation}
where the terms $\lambda_i \geq 0$ describe the Lagrangian multipliers, 
and $\bm{\lambda} = (\lambda_1, \ldots, \lambda_m)$ denotes the vector 
of all multipliers associated with the problem constraints.
In this formulation, the expressions $\max(0, g_i(y))$ capture a 
quantification of the constraint violations, which are often exploited in 
augmented Lagrangian methods \cite{Hestenes:69} and constraint programming 
\cite{Fontaine:14}.

%
When using a Lagrangian function, the optimization problem becomes
\begin{equation}
\label{eq:LR}
LR_{\bm{\lambda}} = \argmin_y f_{\bm{\lambda}}(y),
\end{equation}
and it satisfies $f(LR_{\bm{\lambda}}) \leq f({\cal P})$. That is, 
the Lagrangian function is a lower bound for the original function. 
Finally, to obtain the strongest Lagrangian relaxation of ${\cal P}$,
the \emph{Lagrangian dual} can be used to find the best Lagrangian
multipliers,
\begin{equation}
\label{eq:LD}
\textstyle LD = \argmax_{\bm{\lambda} \geq 0} f(LR_{\bm{\lambda}}).
\end{equation}
For various classes of problems, the Lagrangian dual is a strong
approximation of ${\cal P}$. Moreover, its optimal solutions can often
be translated into high-quality feasible solutions by a
post-processing step, which is the subject of Section \ref{sec:projections}.

\subsection{Augmented Lagrangian of the JSP Constraints}
Given an enumeration of the JSP constraints, the  \emph{violation degree} of constraint $i$ is represented by  $\max(0,g_i(\bm{y}))$. Given the predicted 
values $\hat{\bm{s}}$, the violation degrees associated with the JSP 
constraints are expressed as follows:
\begin{subequations}
\begin{flalign}
    \label{eq7a}
     &\nu_{2b}\hugP{\hat{s}_t^j} = 
     \max \hugP{0, \hat{s}_{t}^j + d_{t}^j - \hat{s}_{t+1}^j} \\
     \label{eq7b}
     &\nu_{2c}\hugP{\hat{s}_t^j, \hat{s}_{t'}^{j'}} =
        \min\hugP{\nu_{2c}^L\hugP{\hat{s}_t^j, \hat{s}_{t'}^{j'}}, 
         \nu_{2c}^R\hugP{\hat{s}_t^j, \hat{s}_{t'}^{j'}}},
\end{flalign}
\end{subequations}
for the same indices as in Constraints \eqref{con:2b} and \eqref{con:2c} respectively, where
\begin{align*}
    \nu_{2c}^L\hugP{\hat{s}_t^j, \hat{s}_{t'}^{j'}} &= 
    \max\hugP{0, \hat{s}_t^j + d^t_j - \hat{s}_{t'}^{j'}}\\
        \nu_{2c}^R\hugP{\hat{s}_t^j, \hat{s}_{t'}^{j'}} &= 
    \max\hugP{0, \hat{s}_{t'}^{j'} + d^{t'}_{j'} - \hat{s}_{t}^{j}}.
\end{align*}

\noindent
The term $\nu_{2b}$ refers to the task-precedence 
constraint \eqref{con:2b}, and the violation degree $\nu_{2c}$ refers to the disjunctive non-overlap constraint 
\eqref{con:2c}, with $\nu_{2c}^L$ and $\nu_{2c}^R$ referring to the two 
disjunctive components. 
When two tasks indexed $(j,t)$ and $(j',t')$ are scheduled on the same 
machine, if both disjunctions are violated, the overall violation degree 
$\nu_{2c}$ is considered to be the smaller of the two degrees, since this is the 
minimum distance that a task must be moved to restore feasibility.

\subsection{The Learning Loss Function}
The loss function of the learning model used to train the proposed 
JSP-DNN can now be augmented with the Lagrangian terms and expressed 
as follows:
\begin{equation}
{\cal L}(\bm{s}, \hat{\bm{s}}, \bm{d}) = {\cal L}(\bm{s}, \hat{\bm{s}})
+ \sum_{c \in {\cal C}} \lambda_c \nu_c(\hat{\bm{s}}, \bm{d}).
\end{equation}
It minimizes the prediction loss--defined as mean squared error between the optimal start times $\bm{s}$ 
and the predicted ones $\hat{\bm{s}}$--and it includes the Lagrangian relaxation based on the violation degrees 
 $\nu$ of the JSP constraints $c \in {\cal C}$.

\section{The Learning Model}
Let $\hat{\cal P}_{\bm{\theta}}$ be the resulting \jssdnn{} with parameters 
$\bm{\theta}$ and let ${\cal L}[\bm{\lambda}]$ be the loss function parametrized by the
Lagrangian multipliers $\bm{\lambda} = \{\lambda_c\}_{c \in {\cal C}}$.
The core training aims at finding the weights $\bm{\theta}$ that minimize the loss
function for a given set of Lagrangian multipliers, i.e., it computes
\[
{\it LR}_{\bm{\lambda}} = \min_{\bm{\theta}} {\cal L}[\bm{\lambda}]\left(\bm{s},\hat{\cal P}_{\btheta}(\bm{d})\right).
\]
In addition, \jssdnn{} exploits Lagrangian duality to obtain the optimal Lagrangian multipliers, i.e., it solves
\[
{\it LD} = \max_{\bm{\lambda}} {\it LR}_{\bm{\lambda}}.
\]
The Lagrangian dual is solved through a subgradient method that 
computes a sequence of multipliers
$\bm{\lambda}^1,\ldots,\bm{\lambda}^k,\ldots$ by solving a sequence of
trainings ${\it LR}_{\bm{\lambda}}^0 ,\ldots,{\it
LR}_{\bm{\lambda}}^{k-1},\ldots$ and adjusting the multipliers using the
violations,
\begin{align}
\bm{\theta}^{k+1} &= \argmin_{\bm{\theta}} {\cal L}[\bm{\lambda}^k]\left(\bm{s},\hat{\cal P}_{\btheta^k}(\bm{s})\right) \label{eq:L1} \tag{L1}\\
\bm{\lambda}^{k+1} &= \left(\lambda^k_c + \rho\,\nu_c\left(\hat{\cal P}_{\btheta^{k+1}}(\bm{s}), \bm{d}\right) \;|\; c\in {\cal C}\right), \label{eq:L2} \tag{L2}
\end{align}
\noindent
where $\rho > 0$ is the Lagrangian step size.
In the implementation, step \eqref{eq:L1} is approximated using a
Stochastic Gradient Descent (SGD) method. Importantly, this step does not
recompute the training from scratch but uses a warm start for the
model parameters $\bm{\theta}$.

\begin{algorithm}[!t]
  \caption{Learning Step}
  \label{alg:learning}
  \setcounter{AlgoLine}{0}
  \SetKwInOut{Input}{input}

  \Input{${\cal D},\alpha, \rho:$ Training data, Optimizer, and Lagrangian step sizes, reps.\!\!\!\!\!\!\!\!\!\!}
  \label{line:1}
  $\bm{\lambda}^0 \gets 0 \;\; \forall c \in {\cal C}$\\
  \For{epoch $k = 0, 1, \ldots$} { 
  \label{line:2}
    \ForEach{$(\bm{d}, \bm{s}) \!\gets\! \mbox{minibatch}({\cal D})$ of size $b$}{
      \label{line:3}
        $\hat{\bm{s}} \gets \hat{\cal P}_{\bm{\theta}}(\bm{d})$\\
        \label{line:4}
        ${\cal L}(\bm{s}, \hat{\bm{s}}) \gets \frac{1}{b}
            \sum_{i \in [b]} 
            {\cal L}\hugP{\bm{s}_i, \hat{\bm{s}}_i} + 
            \sum_{c \in {\cal C}} \lambda_c^k \nu_c \hugP{\bm{d}_i,\hat{\bm{s}}_i}$\!\!\!\!\!\!\!\!\!\!\\
        \label{line:5}
        $\theta \gets \theta - \alpha \nabla_{\theta} 
            \left({\cal L}(\bm{s}, \hat{\bm{s}})\right)$
        \label{line:6}
    }
    \ForEach{$c \in {\cal C}$} {
    \label{line:7}
        $\lambda^{k+1}_c \gets \lambda^k + \rho \nu_c \hugP{\bm{d},\hat{\bm{s}}}$ 
        \label{line:8}
    }
  }
\end{algorithm}

The overall training scheme is presented in
Algorithm \ref{alg:learning}.  It takes as input the training dataset
${\cal D}$, the optimizer step size $\alpha > 0$ and the
Lagrangian step size $\rho > 0$.  The Lagrangian multipliers are
initialized in line \ref{line:1}. The training is performed for a
fixed number of epochs, and each epoch optimizes the weights using a 
minibatch of size $b$. 
After predicting the task start times  (line \ref{line:4}), it 
computes the objective and constraint losses (line \ref{line:5}).  
The latter uses the Lagrangian multipliers $\bm{\lambda}^k$ associated with current
epoch $k$. The model weights $\theta$ are updated in line \ref{line:6}.
Finally, after each epoch, the Lagrangian multipliers are updated
following step \eqref{eq:L2} described above (lines \ref{line:7}
and \ref{line:8}).

\section{The JM-structured Network Architecture}
\label{sec:JM_architecture}

\begin{figure}[!t]
  \centering
  \includegraphics[width=0.85\linewidth]{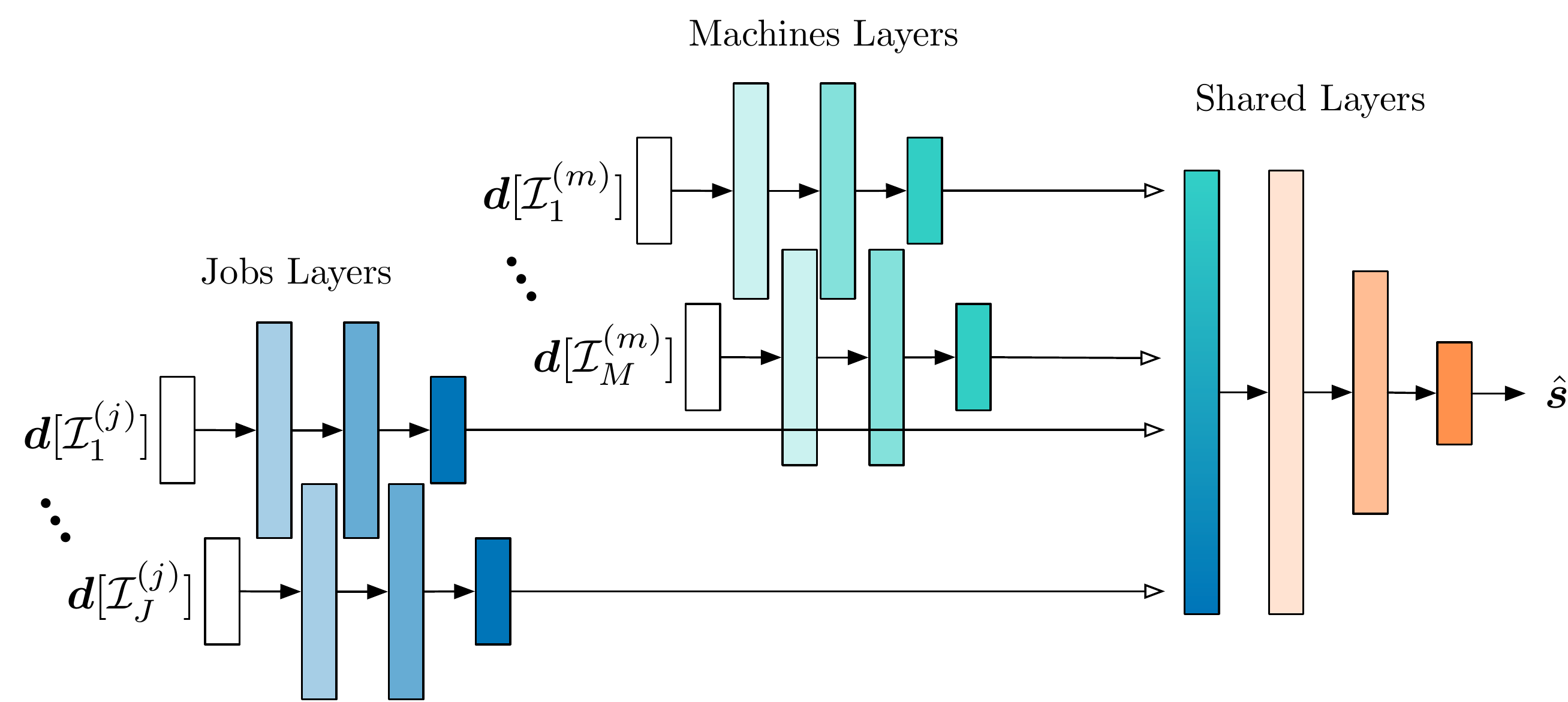}
  \caption{The JM Network Architecture of \jssdnn{}.}
    \label{fig:net_arch}
\end{figure}

This section describes the bespoke neural network architecture that exploits the JSP 
structure. Let $\cI^{(m)}_k$ and $\cI^{(j)}_k$ denote, respectively, the set of 
task indexes associated with the $k^\text{th}$ machine and the $k^\text{th}$ job. Further, denote with $\bm{d}[\cI]$ the set of processing 
times associated with the tasks identified by index set $\cI$. 
The JSP-structured architecture, called JM as for Jobs-Machines, is outlined in Figure \ref{fig:net_arch}. The network differentiates three types of layers: 
\emph{Job layers}, that process processing times organized by jobs, 
\emph{Machine layers}, that process processing times organized by machines,
and \emph{Shared layers}, that process the outputs of the job layers and 
the machine layers to return a prediction $\hat{\bm{s}}$.

The input layers are depicted with white rectangles. Each job layer $k$ takes as input the task processing times $\bm{d}[\cI^{(j)}_k]$ associated with an index set $\cI^{(j)}_k$  $(k \in [J])$, and each machine layer $k$ takes as input the task processing times $\bm{d}[\cI^{(m)}_k]$ associated with an index set $\cI^{(m)}_k$ $(k \in [J])$. The shared layers combine the latent outputs of the job and machine layers. A decoder-encoder pattern follows for each individual group of layers, which are fully connected and use ReLU activation functions (additional details are reported in Section \ref{sec:data_gen}).

\begin{wrapfigure}[17]{r}{200pt}
\vspace{-12pt}
    \centering
    \includegraphics[width=\linewidth]{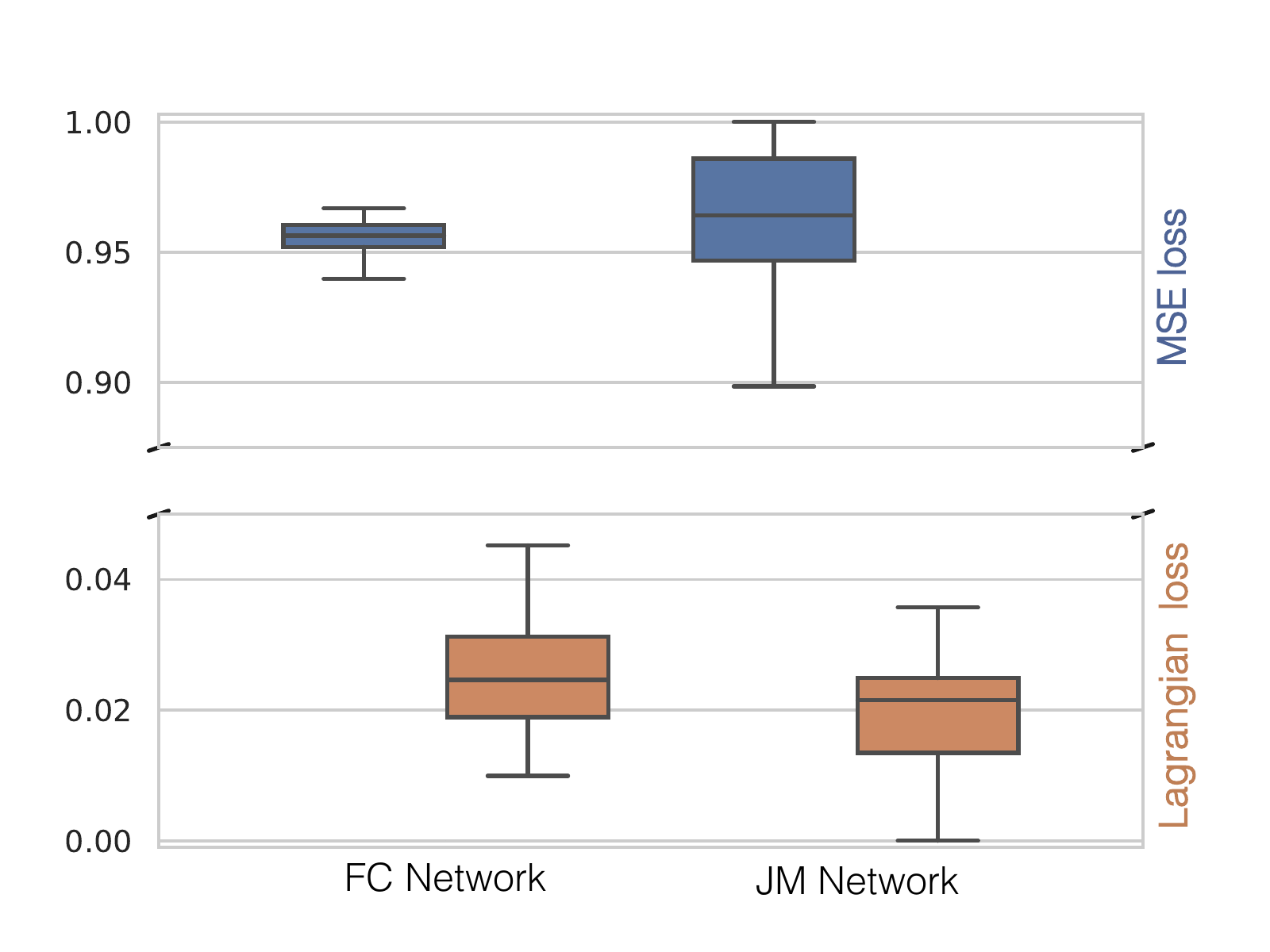}
    \caption{Distribution of No-Overlap Violation results corresponding 
    to the hyperparameter search of Table \ref{tab:hyper_params}.}
    \label{fig:motivation_3}
\end{wrapfigure}
The effect of the JM architecture is twofold. First, the patterns 
created by the job layers and the machine layers reflect the task-precedence 
and no-overlap structure of the JSP, respectively, encouraging
the prediction to account for these constraints.  
Second, in comparison to an architecture that takes as input the full 
vector $\bm{d}$ and process it using fully connected ReLU layers, 
the proposed structure reduces the number of trainable parameters, 
as no connection is created within any two hidden layers processing tasks
on different machines (machine layers) or jobs (job layers). This 
allows for faster convergence to high-quality predictions.

Figure \ref{fig:motivation_3} compares the no-overlap constraint
violations resulting from each choice of loss function and network
architecture over the whole range of hyperparameters searched. 
It uses benchmark \textsl{ta30} and compares the Lagrangian dual results 
(orange/bottom) against a baseline using MSE loss (blue/top). 
It additionally compares the JM network architecture (right) 
against a baseline fully connected (FC) network (left). The results 
are scaled between $0$ (min) and $1$ (max). The figure clearly illustrates 
the benefits of using the Lagrangian dual method, and that the best results 
are obtained when the Lagrangian
dual method is used in conjunction with the JM architecture. The
results for task precedence constraints are analogous.

\section{Constructing Feasible Solutions}
\label{sec:projections}

It remains to show how to recover a feasible solution from the prediction. The recovery method exploits the fact that, given a task ordering on each machine, it is possible to construct a feasible schedule in low polynomial time. Moreover, a prediction $\hat{\bm{s}}$ defines implicitly a task ordering. Indeed, using $(j,t)$ 
to denote task $t$ of job $j$, a start time prediction vector $\hat{\bm{s}}$
can be used to define a task ordering $\preceq$ between tasks executing on the same machine, i.e., 
\[
(j,t) \preceq_{\hat{\bm{s}}} (j', t') \;\;\text{iff}\;\; 
\hat{\mu}_t^j \leq \hat{\mu}_{t'}^{j'} \;\land\; \sigma_t^j = \sigma_{t'}^{j'},
\]
where $\hat{\mu}_{t}^j = \frac{\hat{s}_t^j + d^j_t}{2}$ is the predicted 
midpoint of a task $t$ of job $j$. 
\begin{model}[!t]
{\small
    \caption{Recovering a Feasible Solution to the JSP.}
    \label{model:jss_rec}
    \vspace{-6pt}
    \begin{subequations}
    \begin{align}
    \Pi(\bm{s}) &= {\argmin}_{\bm{s}}\;\;  u \notag \\
        & \mbox{subject to:}\;\;  
            \eqref{con:2a}, \eqref{con:2b}\;\; \notag\\
            & s_t^{j} \geq s_{t'}^{j'} + {d}_{t'}^{j'} 
            \;\; \forall j, j' \!\in\! [J], t,t' \!\in\! [T] \; \text{s.t.} \; (j,t) \preceq_{\hat{\bm{s}}} (j', t') 
            \label{con:11a} \\
            & s_{t}^{j} \geq 0           \;\; \forall j \!\in\! [J], t \!\in\! [T] \label{con:11b}
    \end{align}
    \end{subequations}
    }
    \vspace{-12pt}
\end{model}

The Linear Program (LP) described in Model \ref{model:jss_rec} computes the optimal schedule subject to the ordering $\preceq_{\hat{\bm{s}}}$ associated with prediction $\hat{\bm{s}}$. This LP has the same objective as the original scheduling problem, 
along with the upper bound on start times and task-precedence constraints. 
The disjunctive constraints of the JSP however are replaced by additional 
precedence constraints \eqref{con:11a} and \eqref{con:11b} for the ordering on each machine. The problem is totally unimodular when the durations are integral.

\begin{algorithm}[!t]
{
    \SetAlgoLined
    \setcounter{AlgoLine}{0}
    \KwIn{$\{\hat{s}_t^j\}_{t \in [T], j \in [J]}$: predicted start times}
    $Q \gets \text{enqueue}((j, 1))$, $\;\;\forall j \in [J]$\\
    \While{$Q$ is empty} {
        $(j, t) \gets \text{dequeue}(Q)$ \\
        Schedule task $(j, t)$ with start time $\hat{s}^j_t$\\
        \If{$t < T$} {
            $\hat{s}_{t+1}^{j} \gets \max(\hat{s}_{t+1}^{j},\hat{s}_t^j + d_t^j)$ \\
            $Q \gets \text{enqueue}(j, t)$
        }
        \ForEach{ $(t', j') \text{ s.t. } t \neq t' \; \land \; \sigma_{t'}^{j'} = \sigma_{t}^{j} \; \land \; \hat{s}_{t'}^{j'} \geq \hat{s}_{t}^{j}$} {
            $\hat{s}_{t'}^{j'} \gets  \max(\hat{s}_{t'}^{j'},\hat{s}_t^j + d_t^j)$
        }
    }
    \caption{The JSP Greedy Recovery.}
    \label{alg:greedy}
}
\end{algorithm}

Note that the above LP may be infeasible: the machine precedence constraints may not be consistent with the job precedence constraints. If this happens, it is possible to use a greedy recovery algorithm that selects the next task to schedule based on its predicted starting time and updates predictions with precedence and machine constraints. The greedy procedure is illustrated in Algorithm \ref{alg:greedy} and is organized around a priority queue which is initialized with the first task in each job. Each iteration selects the task with the smallest starting time $\hat{s}_t^j$ and updates the starting of its job successor (line 6) and the starting times of the unscheduled task using the same machine (line 10). It is important to note that this greedy procedure was never needed: \emph{all test cases} (e.g., all 
instances under all hyper-parameter combinations) induced machine orderings that were consistent with the job precedence constraints.

\paragraph{Remark on Learning Task Orderings} 
\jssdnn{} learns to predict schedules as assignments of start times to each task. Another option would have been to learn machine orderings directly in the form of permutations since, once machine orderings are available, it is easy to recover start times. However, learning permutations was found to be much more challenging due to its less efficient representation of solutions. For example, for a $50 \times 10$ JSP, predicting task ordering (or, equivalently, rankings) requires an output dimension of $50 \times 50 \times 10$ (a one-hot encoding to rank each of the $50$
tasks and $10$ machines). In contrast, the start time prediction
requires only one value for each start-time, in this case, $50 \times 10$.
Enforcing feasibility also becomes more challenging in the case of rankings. It nontrivial to design successful violation penalties for task precedence constraints when using the one-hot encoding representation of the solutions.

Start times are easier to predict and indirectly produce good proxies for machine orderings.

\section{Experimental Results}
\label{sec:experiments}

\begin{table}[!tb]
\centering
\resizebox{0.8\linewidth}{!}
{
\begin{tabular}{r l l | r l l }
\toprule
  Parameter & Min Value& Max Value & Parameter & Min Value& Max Value
\\
\midrule                                                                
Learning Rate       $\alpha$&   0.000125  & 0.002     & \texttt{\#} Shared Layers   &   $2$       & $2$   \\
Dual Learning Rate  $\rho$  &   0.001     & 0.05    & Machine Layer Size          &   $2J$      & $2J$    \\
\texttt{\#} Machine Layers  &   $2$       & $2$       & Job Layer Size              &   $2M$      & $2M$    \\
\texttt{\#} Job Layers      &   $2$       & $2$     & Shared Layer Size           &   $2JM$     & $2JM$  \\
\bottomrule
\end{tabular}
}
  \caption{Model Parameters and Hyper-Parameters. 
  \label{tab:hyper_params}}
\end{table}

\begin{table*}[!tb]
\centering
\resizebox{1.0\linewidth}{!}
{
\begin{tabular}{rl | ll ll ll lllll || ll}
\toprule
Instance & \multicolumn{1}{c}{Size} 
         & \multicolumn{2}{c}{Prediction Err($\times 10$) $\downarrow$}
         & \multicolumn{2}{c}{Constraint~Viol($\times 10^2$) $\downarrow$}
         & \multicolumn{5}{c}{Opt.~Gap Heuristics (\%) $\downarrow$}
         & \multicolumn{2}{c||}{Opt.~Gap DNNs (\%) $\downarrow$}
         & \multicolumn{2}{c}{Time SoTA Eq.~(s) $\uparrow$} \\
\cmidrule(r){3-4} 
\cmidrule(r){5-6}
\cmidrule(r){7-11}
\cmidrule(r){12-13} 
\cmidrule(r){14-15} 
  & $J\times M$
  & FC & JSP-DNN
  & FC & JSP-DNN
  & SPT & LWR & MWR & LOR & MOR
  & FC & JSP-DNN
  & FC & JSP-DNN \\
\cmidrule(r){1-13}       
\cmidrule(r){14-15}                                                             
yn02   & $20 \!\times\! 20$ & 2.770 & \textbf{0.138} & 1.134 & \textbf{0.122} & 628 & 837 & 40 & 934 & 40  & 12.80 $\pm$ 5.4 & \textbf{-0.045} $\pm$ \textbf{0.9}  & 10.20 & \textbf{1800+} \\ 
ta25   & $20 \!\times\! 20$ & 1.607 & \textbf{0.361} & 0.631 & \textbf{0.244} & 593 & 877 & 59 & 787 & 46  & 13.61 $\pm$ 3.13 & \textbf{-0.143} $\pm$ \textbf{0.8}  & 11.02 & \textbf{1800+} \\ 
ta30   & $30 \!\times\! 15$ & 4.338 & \textbf{1.196} & 1.483 & \textbf{0.357} & 558 & 910 & 63 & 856 & 46  & 15.01 $\pm$ 2.63 & \textbf{-0.48} $\pm$ \textbf{5.18}  & 9.06  & \textbf{1800+} \\ 
ta40   & $30 \!\times\! 20$ & 7.880 & \textbf{3.341} & 1.863 & \textbf{0.104} & 492 & 794 & 57 & 836 & 25  & 23.11 $\pm$ 7.33 & \textbf{3.19} $\pm$ \textbf{1.88}   & 8.40  & \textbf{12.04} \\ 
ta50   & $50 \!\times\! 10$ & 4.580 & \textbf{1.322} & 1.223 & \textbf{0.225} & 789 & 789 & 53 & 1116 & 43 & 18.30 $\pm$ 5.22 & \textbf{5.85}$\pm$ \textbf{2.72}    & 8.02  & \textbf{90.30} \\ 
swv03  & $20 \!\times\! 15$ & 9.473 & \textbf{2.683} & 2.777 & \textbf{0.850} & 203 & 212 & 75 & 190 & 50  & 28.61 $\pm$ 14.27 & \textbf{7.62} $\pm$ \textbf{2.51}    & 4.04  & \textbf{36.36} \\ 
swv05  & $20 \!\times\! 10$ & 6.586 & \textbf{2.950} & 2.325 & \textbf{0.626} & 183 & 192 & 80 & 177 & 66  & 20.78 $\pm$ 10.54 & \textbf{6.34} $\pm$ \textbf{1.82}   & 7.24  & \textbf{18.18} \\ 
swv07  & $20 \!\times\! 10$ & 4.587 & \textbf{0.681} & 1.222 & \textbf{0.223} & 299 & 295 & 68 & 352 & 43  & 10.69 $\pm$ 6.83 & \textbf{0.01}$\pm$ \textbf{4.75}   & 26.0  & \textbf{254.5} \\ 
swv09  & $20 \!\times\! 15$ & 5.678 & \textbf{3.462} & 2.132 & \textbf{0.211} & 322 & 270 & 69 & 285 & 75  & 22.12 $\pm$ 8.52 & \textbf{5.42} $\pm$ \textbf{1.21}   & 6.48  & \textbf{28.32} \\ 
swv11  & $50 \!\times\! 10$ & 7.958 & \textbf{3.244} & 2.711 & \textbf{0.282} & 237 & 231 & 94 & 263 & 73  & 23.18 $\pm$ 2.27 & \textbf{4.80} $\pm$ \textbf{4.47}     & 7.02  & \textbf{92.00} \\ 
swv13  & $50 \!\times\! 10$ & 23.21 & \textbf{3.557} & 1.615 & \textbf{0.323} & 225 & 203 & 114 & 218 & 79 & 22.79 $\pm$ 16.21 & \textbf{8.11} $\pm$ \textbf{4.20}   & 7.08  & \textbf{24.08} \\ 
\bottomrule
\end{tabular}
}
  \caption{Accuracy metrics compared between FC and JSP-DNN (left sub-table) 
  and accuracy of simple heuristics vs CP-Optimizer at 1800s (right sub-table). 
  Best results shown in bold. }
    \label{tab:JSP_acc}
\end{table*}

The experiments evaluate \jssdnn{} against the baseline FC network, 
the state-of-the-art CP commercial solver IBM CP-Optimizer, and several well-known 
scheduling heuristics. 

\paragraph{Data Generation and Model Details}
\label{sec:data_gen}

A single instance of the JSP with $J$ jobs and $M$ machines is defined by a set of
assignments of tasks to machines, along with processing times
required by each task.  The generation of JSP instances
simulates a situation in which a scheduling system experiences an
unexpected ``slowdown'' on some arbitrary machine, inducing an 
increase in the processing times of each task assigned to the
impaired machine. To create each  experimental dataset, a root
instance is chosen from the JSPLIB repository 
\cite{tamy0612_2014}, and a set of $5000$ individual problem instances are
generated accordingly, with all processing times on the slowdown machines
extended from their original value to a maximum increase of $50$
percent. Each such instance is solved using the IBM CP-Optimizer constraint-programming 
software \cite{laborie2009ibm} with a time limit of 1800s. The 
instance and its solution are included in the training dataset.

\paragraph{Model Configuration}
\label{sec:data_gen}

To ensure a fair analysis, the learning models have the same 
number of trainable parameters. The JM-structured neural networks 
are given two decoder layers per
job and machine, whose sizes are twice as large as their
corresponding numbers of tasks. 
These decoders are connected to a single \emph{Shared Layer} 
(of size $2 \times J \times M$). A final
output layer, representing task start times, has size equal to the
number of tasks in the JSP instance. The baseline fully connected
networks are given $3$ hidden layers, whose sizes are chosen to match the
the size of the corresponding \jssdnn{} network in terms of the total number 
of parameters. The experiments apply the recovery operators discussed in the previous 
section to both the FC and \jssdnn{} models to obtain feasible solutions from their predictions. 

Time required for training and for running the models are reported in Appendix \ref{app:timing}.

The learning models are configured by a hyper-parameter search for the values
summarized in Table \ref{tab:hyper_params}. The search samples evenly spaced learning rates between
their minimum and maximum values, along with similarly chosen dual
learning rates $\rho$ when the Lagrangian loss function is used.
Additionally, each configuration is evaluated with $5$ distinct random seeds,
which primarily influences the neural network parameter initialization. 
The model with the best predicted schedules in average is selected for the
experimental evaluation.

\subsection{Model Accuracy and Constraint Violations}
Table \ref{tab:JSP_acc} represents the performance of the selected models.
Each reported performance metric is averaged over the entire test set. 
For each prediction metric, symbols $\downarrow$ and $\uparrow$ indicates 
whether lower or higher values are better, 
respectively. The evaluation uses a $80$:$20$ split with the training data. 

\smallskip\noindent $\bullet$ \textbf{Prediction Error $\downarrow$}
The columns for prediction errors report the error between a predicted schedule 
and its target solution (before recovering a feasible solution). It is measured 
as the L1 distance between respective start times. 
The predictions by \jssdnn{} are much 
 closer to the targets than those of the FC model. \jssdnn{} 
 reduces the errors by up to an order of magnitude
 compared to FC, demonstrating the ability of the \textsl{JM} architecture and the Lagrangian dual method 
 to exploit the problem structure to derive good predictions. 

\smallskip\noindent $\bullet$ \textbf{Constraint Violation $\downarrow$} 
The constraint violations are collected before recovering a feasible
 solution and the columns report the average magnitude of overlap between two tasks as
 a fraction of the average processing time. The results show again that the
 violation magnitudes reported by \jssdnn{} are typically one order of magnitude lower 
 than those reported by FC. They highlight the effectiveness of the Lagrangian dual approach 
 to take into account the problem constraints. 
 
\smallskip\noindent $\bullet$ \textbf{Optimality Gap $\downarrow$} 
The quality of the predictions is measured as the average relative difference between the makespan of the \emph{feasible solutions} recovered from the predictions of the deep-learning models, and the makespan obtained by the IBM CP-Optimizer with a timeout limit of $1800$ seconds. The optimality gap is the primary measure of solution quality, as the goal is to predict solutions to JSP instances that are as close to optimal as possible. The table also reports the optimality gaps  achieved by several (fast) heuristics, relative to the same CP-Optimizer baseline. They are Shortest Processing Time (SPT), Least Work Remaining (LWR), Most Work Remaining (MWR), Least Operations Remaining (LOR), and Most Operations Remaining (MOR). Since the CP solver cannot typically find optimal solutions within the given timeout, the results under-approximate the true optimality gaps. 
 
The results show that the proposed \jssdnn{}  substantially outperforms all other heuristic baselines and the best FC network of equal capacity. On these difficult instances, the most accurate heuristics exhibit results whose relative errors are at least an order of magnitude larger than those of \jssdnn{}. Similarly, \jssdnn{} significantly outperforms FC. 

It is interesting to observe that, for instances \textsl{yn02, ta25}, and \textsl{ta30}, the makespan of the recovered \jssdnn{} solutions outperform, on average, those reported by the CP solver before it reaches its time limit, which is quite remarkable.
The CP solver can actually produce these solutions as well if it is allowed to run for several hours. This opens an interesting avenue beyond for further research: the use of \jssdnn{} to hot-start a CP solver. 
 
\begin{figure*}[!t]
    \centering
    \includegraphics[width=0.38\linewidth]{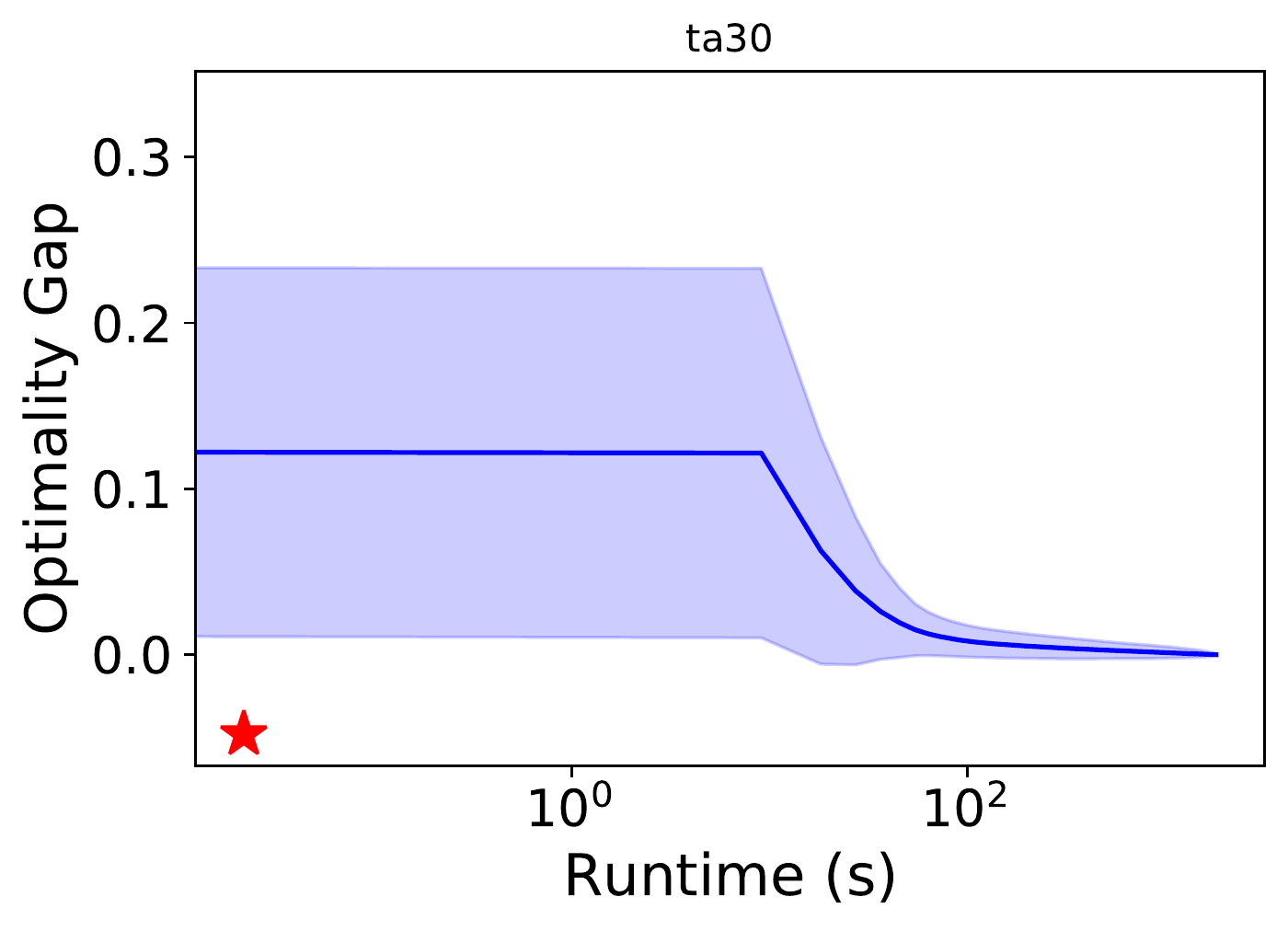}
    \includegraphics[width=0.38\linewidth]{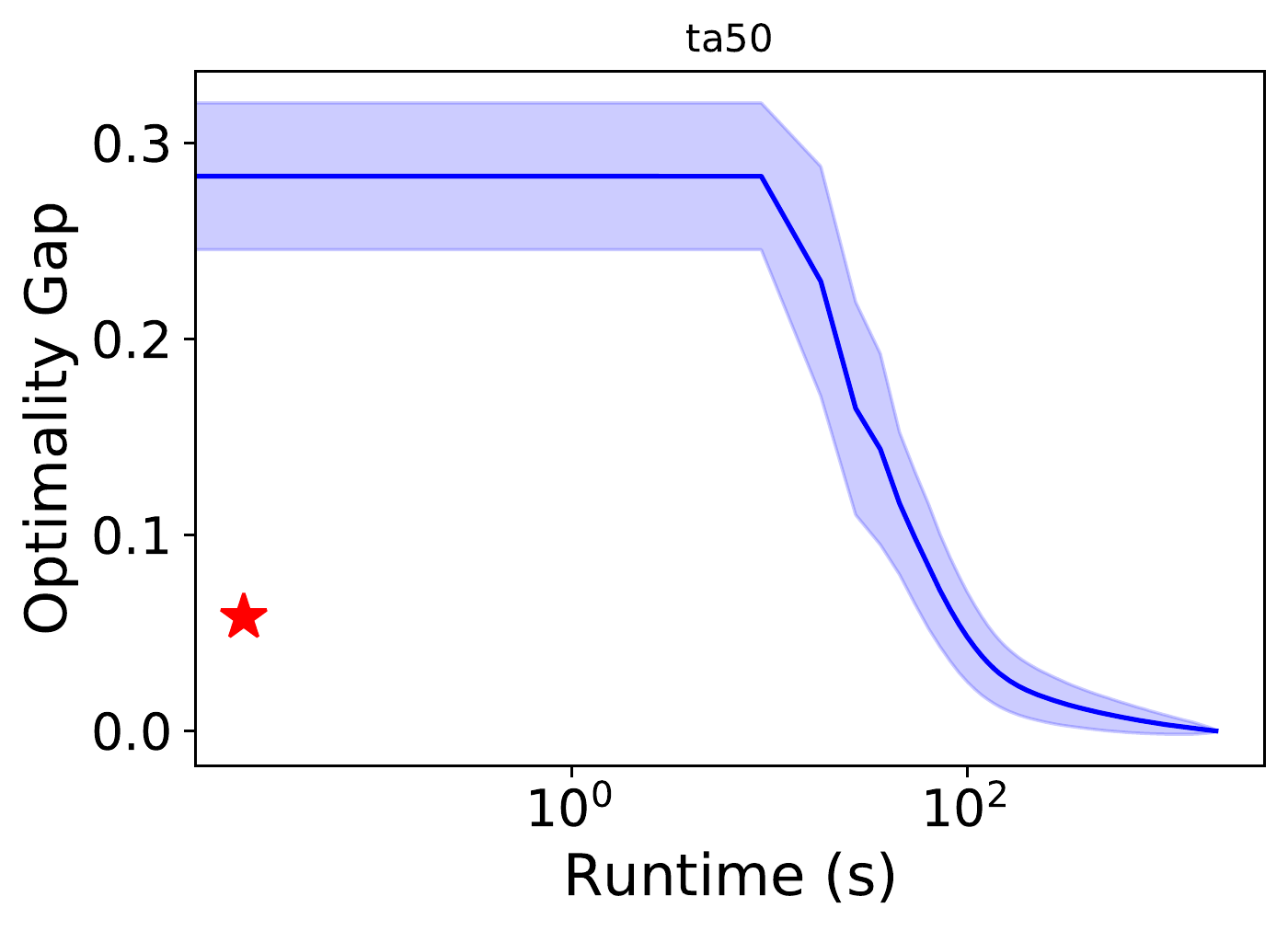}
    \includegraphics[width=0.38\linewidth]{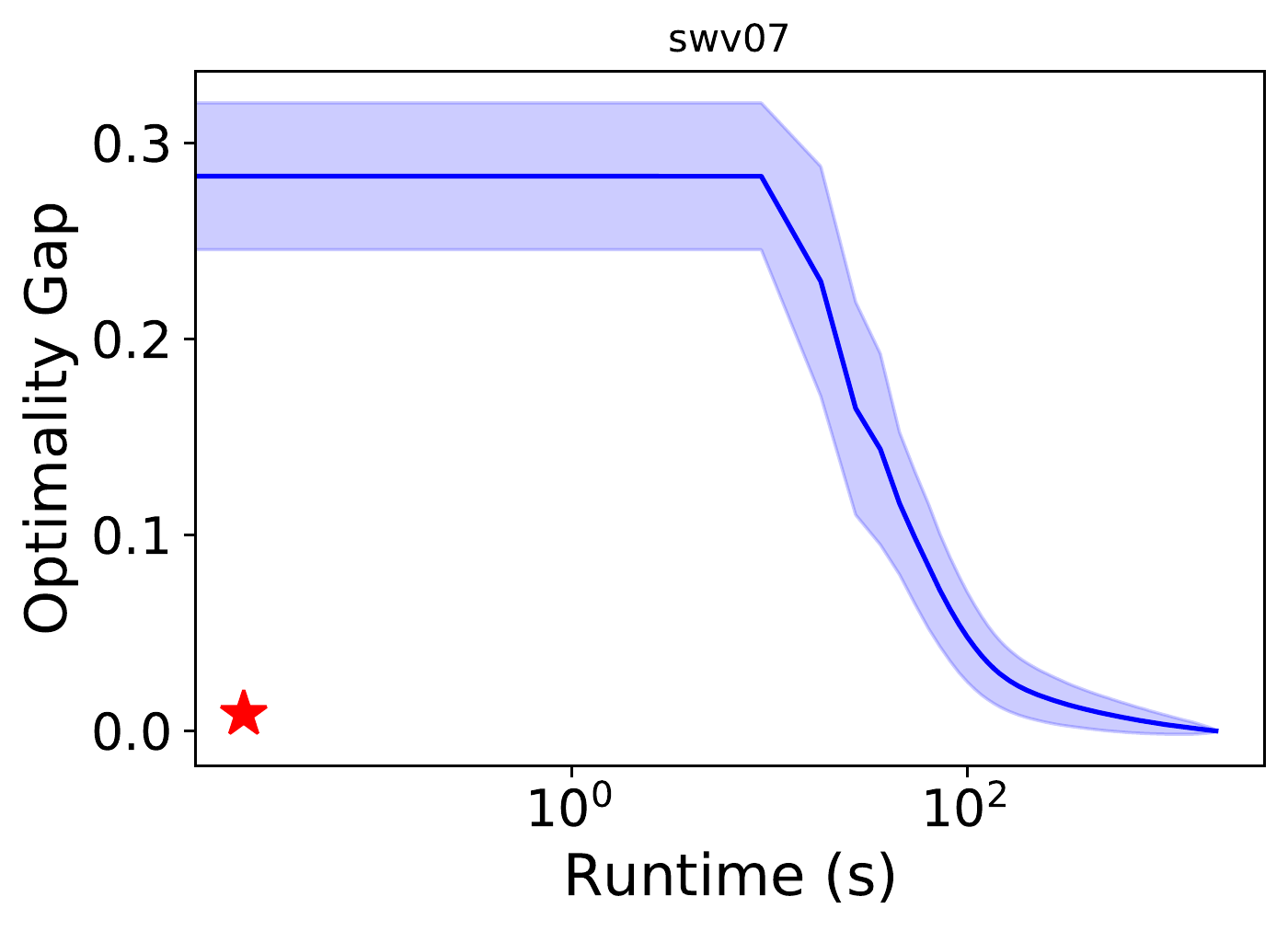}
    \includegraphics[width=0.38\linewidth]{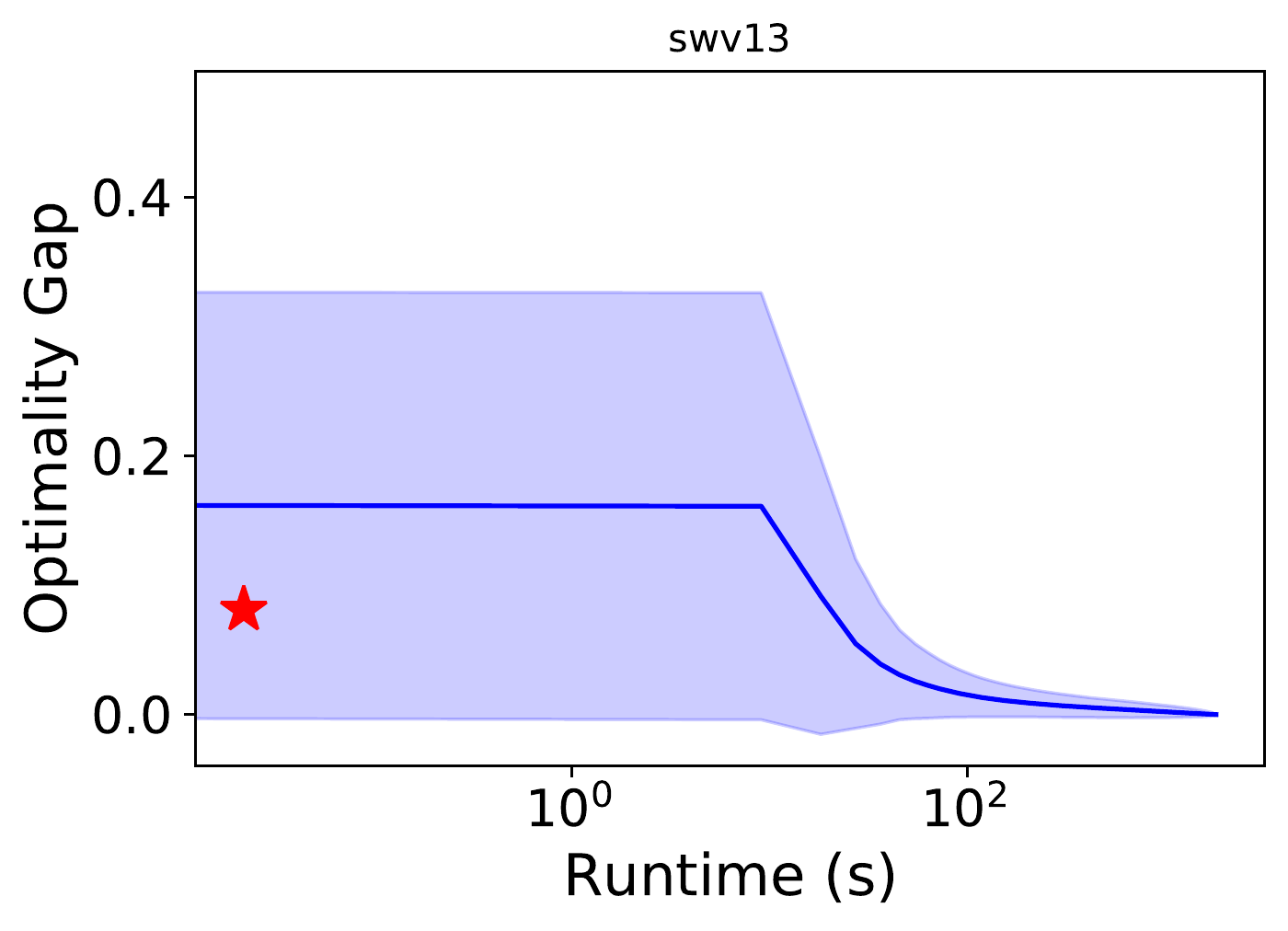}
    \caption{Comparison of mean timing and accuracy of four \jssdnn{} models (red)  against CP-optimizer mean optima and standard deviation (blue).}
    \label{fig:timing_CP}
\end{figure*}

\subsection{Comparison with SoTA Solver $\uparrow$}

The last results compare \jssdnn{} and the state-of-the art CP solver in terms of their ability to find high-quality solutions quickly. The runtime of \jssdnn{} is dominated by the solution recovery process (Model \ref{model:jss_rec}  and Algorithm \ref{alg:greedy}): their runtimes depend on the instance size but never exceed $30$ms for the test cases. 

The results are depicted in the last two columns of Table \ref{tab:JSP_acc}. They report the average runtimes required by the CP solver to produce solutions that match or outperform the feasible (recovered) solutions produced by \jssdnn{} and FC. The results show that it takes the CP solver less than 12 seconds to outperform FC. In contrast, the CP solver takes at least an order of magnitude longer to outperform \jssdnn{} and is not able to do so within 30 minutes on the first 3 test cases. 

Figure \ref{fig:timing_CP} complements these results on three test cases: they depict the evolution of the makespan produced by the CP solver over time and contrast these with the solution recovered by \jssdnn{}. The thick line depicts the mean makespan,
the shaded region captures its standard deviation, and the red star reports the quality of the \jssdnn{} solution. These results highlight the ability of \jssdnn{} to generate a high-quality solution quickly. 

\smallskip
\emph{Overall, these results show that \jssdnn{} may be a useful addition to the set of optimization tools for scheduling applications that require high-quality solutions in real time.} It may also provide an interesting avenue to seed optimization solvers, as mentioned earlier.

\section{Conclusions}
This paper proposed \jssdnn, a deep-learning approach to produce 
high-quality approximations of Job shop Scheduling Problems (JSPs) in 
milliseconds. 
The proposed approach combines deep learning and Lagrangian 
duality to model the combinatorial constraints of the JSP. 
It further exploits the JSP structure through the design of dedicated neural network architectures to reflect the nature of the task-precedence and no-overlap structure of the JSP, encouraging the 
predictions to take account of these constraints. 
The paper also presented efficient recovery techniques to
post-process \jssdnn{} predictions and produce feasible solutions.
The experimental analysis showed that \jssdnn{} 
produces feasible solutions whose quality is at least an order
of magnitude better than commonly used heuristics. 
Moreover, a state-of-the-art commercial CP solver was shown to take a significant amount of time to obtain solutions of the same quality and may not be able to do so within 30 minutes on some test cases.

\section*{Acknowledgement}
This research is partially supported by NSF grant 2007164. Its views and
conclusions are those of the authors only.

\bibliographystyle{abbrvnat}
\bibliography{bib}

\appendix

\pagenumbering{arabic}
\renewcommand{\thepage} {A--\arabic{page}}

\section{Timing Results}
\label{app:timing}

In the models described in this paper, the size of each DNN layer depends on the size of the corresponding benchmark JSP instance, as do the number of constraints to be enforced. These factors influence the computational runtime of each component model and the associated training. Table \ref{tab:timing} presents the time required to train for $1$ epoch, the total training time taken to train each model, the time required to run inference on $1$ sample, and the time required to construct feasible solutions using Model \ref{model:jss_rec}. All results are reported on average.

\begin{table*}[!b]
\centering
\resizebox{.75\linewidth}{!}
{
\begin{tabular}{r l l l l}
\toprule
  Benchmark 
         & Training ($1$ Epoch) (s)
         & Training (Total) (s)
         & Inference (s) ($\times 10^3$)
         & Model \ref{model:jss_rec} (s)
\\
\midrule                                                                
yn02  &   2.306  & 1153 & 3.7  & 0.0152      \\
ta25  &   2.102  &  1051 & 3.7  & 0.0155     \\
ta30  &   2.232  &   1116 &  5.0 & 0.0160      \\ 
ta40  &   2.914 &  1457  & 4.2 & 0.0171     \\
ta50  &   4.452  & 2226  & 7.5 & 0.0301      \\
swv03 &   1.377  & 688 & 2.9  & 0.0095    \\
swv05 &   1.335  & 668 & 3.5  & 0.0086    \\
swv07 &   2.135  & 1067 & 3.1  & 0.0122    \\
swv09 &   2.014  & 1007 & 3.1   & 0.0121    \\
swv11 &   5.018  & 2509 & 5.3 & 0.0213    \\
swv13 &   5.233  & 2617 & 5.4 & 0.0215    \\

\bottomrule
\end{tabular}
}
  \caption{Runtime for Model Training and Execution. 
  \label{tab:timing}}
\end{table*}

\section{Technical Specifications}
\label{app:specs}

All computations involved in this work were performed on the following platform: Intel(R) Xeon(R) Platinum 8260 CPU @ 2.40GHz. The operating system used throughout was Ubuntu 20.04.2 LTS, along with Python $3.7.6$, Pytorch $1.4.0$, Numpy $1.20.3$, Google OR-Tools $ 8.0.8283$, and IBM ILOG CP Optimizer Developer Edition $12.10$.

\section{Code and Data}
\label{app:code_data}

All training datasets consist of $5000$ individual JSP instances,
beginning with a root benchmark instance and with uniformly increased
processing times on one machine, from $1$ to $1.5$ times the original
duration. Each instance is solved using the CP Optimizer software
with a time limit of $30$ minutes. To control for the existence of
symmetries, i.e. co-optimal solutions, each solved instance is
post-processed to minimize the $L^1$ distance from its schedule to
that of the next instance, with respect to increasing processing
times. 
All code used to produce this project will be released upon publication.

\section{Final Hyperparameters}
\label{app:hyper}

Table \ref{tab:hyper} presents the final hyperparameters used for each benchmark instance and model to produce the results of Table \ref{tab:JSP_acc}.

\begin{table*}[!tb]
\centering
\resizebox{.75\linewidth}{!}
{
\begin{tabular}{r l l l l}
\toprule
  Benchmark 
         & Learning Rate (Baseline/FC)
         & Learning Rate (JSP-DNN)
         & Dual Learning Rate (JSP-DNN)
       
\\
\midrule                                                                
yn02  &   0.002667  & 0.009432 & 0.001     \\
ta25  &   0.01  &  0.004141 & 0.001     \\
ta30  &   0.01  &   0.004141 &  0.001     \\ 
ta40  &   0.01 &  0.01  & 0.01   \\
ta50  &   0.000833  & 0.001797  & 0.001      \\
swv03 &   0.0045  & 0.001797 & 0.001     \\
swv05 &   0.0045  & 0.007515 & 0.001     \\
swv07 &   0.000833  & 0.004141 & 0.05   \\
swv09 &   0.01  & 0.001797 & 0.001       \\
swv11 &   0.008167  & 0.002386 & 0.001    \\
swv13 &   0.001  & 0.001797 & 0.001   \\

\bottomrule
\end{tabular}
}
  \caption{Final Hyperprarmeters used. 
  \label{tab:hyper}}
\end{table*}

\end{document}